\documentclass[11pt]{article}
\usepackage{epsfig, subfigure, caption2, latexsym, ifthen, amsfonts, amsbsy}
\usepackage[round,comma]{natbib}

\newcaptionstyle{one}{ % for putting captions below figures
  \usecaptionmargin\captionfont
    \onelinecaption
   {{\small\bfseries\captionlabelfont\captionlabel\captionlabeldelim} {\small\captiontext}}
   {{\centering\small\bfseries\captionlabelfont\captionlabel\captionlabeldelim} {\small\captiontext}}
}

\newcommand{\lhead}{\ensuremath{\prec}}

\newcommand{\head}{\ensuremath{\succ}}

\newcommand{\pedg}[2]{\ensuremath{{\kern0.5pt
\scriptstyle{\ifthenelse{\equal{\head}{#1}}{\lhead}{#1}}\joinrel\relbar
\joinrel{#2}\kern0.5pt}}}
\newcommand{\lra}{\pedg{\head}{\head}}   % <-->
\newcommand{\spo}{{\rm sp}}      % for spouces
\newtheorem{theorem}{Theorem}

\title{Supplementary Material for \\ Markov Equivalence for Ancestral Graphs}

\author{R. Ayesha Ali\\
{\small University of Guelph, Canada}\\
{\texttt{aali@uoguelph.ca}}\\ \and
Thomas Richardson\\ 
{\small University of Washington, USA}\\
\texttt{tsr@stat.washington.edu} \and 
Peter Spirtes\\
{\small Carnegie-Mellon University, USA}\\
\texttt{ps7z@andrew.cmu.edu} 
}

\begin{document}

\maketitle

\begin{abstract}

We prove that the criterion for Markov equivalence provided by
\cite{zhao:markov:2005} may involve a set of features of a graph that is
exponential in the number of vertices.

\end{abstract}

\begin{appendix}

\cite{zhao:markov:2005} define a collider path $\mathbf{\nu}  = \langle v_1, \ldots , v_n \rangle$
to be minimal if there is no subsequence of vertices $ \langle v_1 = v_{i_1} , \ldots , v_{i_k} = v_n \rangle$
which forms a collider path. They use this to provide the following simple characterization
of Markov equivalence for maximal ancestral graphs:

\begin{theorem} {\bf (Zhao, Zheng, Liu)} Two maximal ancestral graphs ${\cal G}_1$ and
${\cal G}_2$ are Markov equivalent if and only if ${\cal G}_1$ and ${\cal G}_2$ have the same 
minimal collider paths.
\end{theorem}

This is a very elegant characterization, but unfortunately the number
of such minimal collider paths may grow exponentially with the number of
vertices, making the criterion computationally infeasible for large graphs:

\begin{figure}[h]
\captionstyle{one}
\begin{center}
\setlength{\unitlength}{0.2in}
\psfig{figure=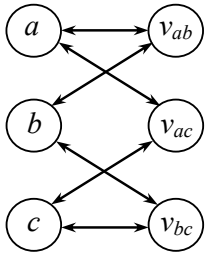,width=1.2in}
\end{center}
\vspace{-15pt}
\caption{The bi-partite bi-directed graph ${\cal G}_3$.}
\label{bipartitie}
\end{figure}

Let ${\cal G}_k$ be a bi-directed graph with vertex set $V = V_1 \dot{\cup} V_2$. Here $|V_1| = k,
V_2 = \{v_{ab} \/\/\/ | \/\/ a, b \in V_1, a \neq b \}$ and edge set $E = \{a \lra v_{ab} | a \in V_1, v_{ab} \in
V_2 \}$; we do not distinguish $v_{ab}$ and $v_{ba}$. Thus ${\cal G}_k$  is a bi-partite graph with
$k(k + 1)/2$ vertices in total. For every pair of vertices $a, b$ in $V_1$, there is
a vertex $v_{ab}$ in $V_2$ with edges $a \lra v_{ab} \lra b$. There are no
edges between vertices in $V_1$, nor between vertices in $V_2$. If $a \in V_1$ then
$|\spo(a)| = k - 1$; if $v_{ab} \in V_2$ then $|\spo(v_{ab})| = 2$. We may associate a unique
path with ordered endpoints with every ordered sequence of vertices in $V_1$
as follows:
\[ \langle a_1,\ldots , a_k \rangle \longmapsto \langle a_1, v_{a_1 a_2} , a_2, \ldots, a_{k-1}, v_{a_{k-1}a_k} , a_k \rangle.\]
Thus there are $k!$ such ordered sequences, hence the number of minimal
collider paths in ${\cal G}_k$ is greater than $k!/2$. To complete the argument we
observe that for a fixed $n$, if we define $k(n) = \max_k k(k + 1)/2 \leq n$, then
$k(n) \geq \sqrt{n}$. Hence ${\cal G}_{k(n)}$ has at most $n$ vertices but $O ((\sqrt{n})!/2)$ minimal
collider paths.

\end{appendix}

\end{document}